\documentclass[preprint]{oscmjournal}

\usepackage{algorithm}
\usepackage{algpseudocode}
\usepackage{amsthm}
\usepackage{graphicx}
\usepackage{booktabs}
\usepackage{amsmath}
\usepackage{amssymb}

\renewcommand{\JournalVolume}{00}
\renewcommand{\JournalNumber}{0}
\renewcommand{\JournalYear}{2026}
\renewcommand{\FirstPage}{1}
\renewcommand{\RunningHeader}{Lastname1 \& Lastname2: Author Guidelines for Preparing Manuscripts for the OSCM Journal}

\title{Drifting Models for Surrogate Flow Modeling}

\author[*]{$^1$Chris R. Jung$^\dag$; $^1$Markus Dörr$^\dag$; $^2$Natalie Jüngling; $^2$Jennifer Niessner; $^1$Adam T. Müller$^{\ddag}\textsuperscript{*}$; $^1$Nicolaj C. Stache$^\ddag$}
  {$^1$Center for Machine Learning (ZML)\\
   $^2$Institute for Flow in Additively Manufactured Porous Structures (ISAPS)\\
   Heilbronn University of Applied Sciences, 74081 Heilbronn, Germany}


\begin{document}

\maketitle



\begin{abstract}

\noindent While Computational Fluid Dynamics (CFD) provides high-fidelity flow fields for optimizing indoor environments, its computational cost limits rapid exploration. To solve this problem generative surrogates offer better distribution modeling than deterministic networks, but iterative sampling is slow. To enable high-quality, single-pass generation, we adapt the novel generative drifting framework to fluid mechanics. We introduce a conditional architecture that performs drifting in a learned VAE latent space and uses label-aware masking to align generated samples with their boundary conditions. Our label-conditioned model matches iterative diffusion in accuracy and flow consistency while running two orders of magnitude faster. Additionally, we propose a spatial-conditioning variant that establishes a promising path towards generalization to unseen geometries. Ultimately, conditional drifting serves as a highly efficient alternative to diffusion based approaches, unlocking real-time CFD surrogates where inference speed is critical.
\end{abstract}

\section{Introduction}\label{sec:intro}

Indoor air quality (IAQ) is critical for human health, since humans typically spend most of their time in indoor environments \cite{concilio_cfd_2024}. Ventilation (e.g. window ventilation or mechanical ventilation) effectively decreases the concentration of all indoor contaminants. Designing them efficiently requires an accurate understanding of complex indoor airflow patterns, which are highly sensitive to room geometry, inlet locations, and flow velocities \cite{lin_cfd_2005, hatif_cfd_2020}.   

While Computational Fluid Dynamics (CFD) provides the high-fidelity flow fields necessary to optimize such environments \cite{topak_collective_2023}, its notable computational cost precludes rapid design exploration and real-time ventilation control \cite{bian_operator_2026, tobisch_reducing_2021}. This creates a need for data-driven surrogate models capable of delivering reliable predictions with reduced computational demands \cite{mao_rapid_2025}.

While deterministic architectures like Fourier Neural Operators (FNOs) \cite{marwah_deep_2023} and coordinate-based networks \cite{muller_reducing_2025} offer rapid inference, capturing environments with complex multimodal flow distributions remains challenging. Generative approaches, such as diffusion models, successfully capture this physical complexity. However, their iterative nature makes inference computationally heavy \cite{luo_difffluid_2024}.

In this work, we explore the potential of adapting novel generative modeling via drifting \cite{deng_generative_2026} to the domain of fluid mechanics. Originally designed to evolve pushforward distributions during training to enable high-quality, one-step generation without iterative inference, we demonstrate an adapted conditional drifting architecture on 2D indoor flow modeling as a foundational proof-of-concept. Our main contributions are as follows: \textbf{(1)} We prove that conditional drifting models can accurately and efficiently generate flow fields in a single step. \textbf{(2)} We introduce essential modifications for physical conditioning, namely latent-space drifting via a learned Variational Autoencoder (VAE) and label-aware positive masking to enforce boundary conditions. \textbf{(3)} We develop a spatial encoding variant to explore generalization across unseen room geometries.

\section{Related Work}\label{sec:relWork}

\textit{Data-Driven Surrogates for Flow Prediction.} Deep learning architectures, including convolutional neural networks (e.g., U-Nets) and FNOs \cite{muller_reducing_2025, li_fourier_2020}, are increasingly used as surrogates of traditional CFD solvers \cite{marwah_deep_2023, muller_reducing_2025}. While deterministic surrogates can offer dramatic inference speedups and establish strong global flow priors, they are mostly designed to minimize standard regression losses like Mean Squared Error. Consequently, such surrogates tend to predict a smoothed expectation of the flow. This can fail to capture the complex, high-frequency, and multimodal distributions characteristic of indoor environments, where minor variations in boundary conditions or obstacles dictate drastically different, yet equally valid, flow regimes \cite{lin_cfd_2005}.

\noindent \textit{Generative Modeling for Physical Dynamics.} To address this smoothing and enable uncertainty-aware predictions, generative architectures have gained traction. Diffusion model based approaches have been established as surrogates for aerodynamic simulations, successfully capturing complex, physically plausible flows~\cite{luo_difffluid_2024}. Advanced autoregressive variants extend these capabilities to temporal predictions~\cite{kohl_benchmarking_2026}. However, diffusion relies on an iterative denoising process. The resulting hundreds of neural function evaluations (NFE) render generation computationally heavy, contradicting the goal of rapid surrogate modeling. Flow matching approaches reduce this burden by learning continuous-time conditional flows, requiring fewer sampling steps while extending to unstructured geometries~\cite{baldan_physics_2025, ramos_fluidflow_2026}. Nevertheless, flow matching still relies on numerically integrating an ordinary differential equation during inference~\cite{lipman_flow_2022}. To bypass such iterative sampling, we build upon the recently introduced framework of generative modeling via drifting~\cite{deng_generative_2026}. Originally designed for image generation and grounded in optimal transport~\cite{he_sinkhorn-drifting_2026}, drifting evolves pushforward distributions during training. By enabling high-quality, one-step generation ($\text{NFE}=1$), it provides a promising foundation for rapid fluid surrogates.

\section{Method}\label{sec:method}

\subsection{Fundamentals of Drifting Models.}

A drifting model \cite{deng_generative_2026} is a single-pass network $f : \mathbb{R}^C \to \mathbb{R}^D$ that maps a prior sample $\epsilon \sim p_\epsilon$ to an output $x = f(\epsilon)$, inducing a pushforward distribution $q = f_\# p_\epsilon$ that is trained to match the data distribution $p$. Rather than learning an iterative noise-to-data trajectory as in diffusion or flow matching, drifting models evolve $q$ at training time through a drifting field $V_{p,q}(x)$ that prescribes, at each iteration, the direction in which a generated sample should move so that $q$ approaches the data distribution $p$. The field is constructed to satisfy $V_{p,q}(x) = 0$ whenever $q = p$, so equilibrium coincides with distribution matching. Training thus reduces to a fixed-point update $f_{\theta_{i+1}}(\epsilon) \leftarrow f_{\theta_i}(\epsilon) + V_{p,q_{\theta_i}}(f_{\theta_i}(\epsilon))$, realised as the stop-gradient loss: 
\begin{equation}
    \mathcal{L} = \mathbb{E}_{\epsilon} \big[\big\| f_\theta(\epsilon) - \mathrm{stopgrad}\big(f_\theta(\epsilon) + V_{p,q_\theta}(f_\theta(\epsilon))\big) \big\|^{2}\big].
\end{equation}
In practice $V_{p,q}$ is instantiated as a kernel-based attraction--repulsion field driven by positive samples drawn from the data and negative samples from $q$, with the negative samples maintained in a memory bank to amortise their cost across iterations. As $f_\theta$ is a single-pass network, inference is one-step (NFE${=}1$), in contrast to the multi-step solvers required by diffusion and flow-based baselines.

\subsection{Adjustments for Surrogate Modelling}
 
We adapt the original drifting model along three axes that reflect the different demands of CFD surrogate modelling compared to natural-image generation: the operating space, the conditioning interface, and the construction of the drifting field.

\noindent \textit{Latent-space drifting via a learned VAE.} Unlike the original work using an ImageNet-pretrained MAE, we trained a convolutional VAE on our simulation data to ensure a domain-specific feature extractor. The VAE maps $64\times64$ velocity fields to a $4\times4\times16$ latent representation $z = E(\mathbf{u})$ with encoder $E$. By performing drifting within this compact, semantically organized latent space and reconstructing via the decoder $D$, we maintain structural integrity while avoiding suboptimal pre-trained features.


\noindent \textit{Conditioning.}
Whereas the original generator is conditioned on a single class label, each of our samples carries a condition comprising the inlet position, the outlet position, the room geometry, and a scalar inlet velocity magnitude $v_\mathrm{in} \in \mathbb{R}$. We investigate two encoder variants that differ in how the geometric information is presented to the network:

\begin{itemize}

\item \textbf{Label-based.} The \emph{label-based} variant treats inlet and outlet as discrete identifiers $(\text{inlet\_id}, \text{outlet\_id})$ $\in \mathbb{N}$, drawn from a fixed catalogue of room configurations and embeds each through a learned embedding table. The room geometry is implicit in the identifier. The label-based variant is restricted to the configurations enumerated in its embedding table.

\item \textbf{Spatial.} The \emph{spatial} variant replaces these identifiers with a three-channel binary mask M with ${3 \times H \times W}$ that explicitly encodes the inlet, outlet, and obstacle locations on the simulation grid, and processes it through a small convolutional encoder that produces a fixed-dimensional vector. The spatial variant is more flexible than the label-based variant and can theoretically generalise better to arbitrary room layouts, including geometries unseen during training.  

\end{itemize}

\noindent In both cases $v_\mathrm{in}$ is embedded by a small Multilayer Perceptron (MLP) and summed with the geometry embedding, where the resulting conditioning vector $c$ drives the adapted Diffusion Transformer (DiT) backbone \cite{peebles_scalable_2023}.

\noindent \textit{Label-aware positive masking.} To adapt the kernelized drifting formulation \cite{deng_generative_2026} to our data-sparse setting, we introduced a binary compatibility mask. Unlike the original work \cite{deng_generative_2026}, which assumes large pools of ground-truth samples per class and processes them in uniform batches, our approach allows mixed conditions within a batch. We gate the attraction kernel by matching $(\text{inlet}, \text{outlet}, \text{geometry})$ triples, ensuring generated samples only drift toward compatible positives. While the global repulsion term remains unchanged, this masking prevents the attraction term from averaging across incompatible flow regimes, which would otherwise force samples toward a condition-agnostic mean.


\subsection{Latent Diffusion Baseline.} \label{chap:diffusion}

We adopted a Latent Diffusion Model baseline \cite{rombach_2021} inspired by Liu and Thuerey’s state-of-the-art work in RANS-based surrogates \cite{thuerey_2023}. Operating in the same $4{\times}4$ latent space as our drifting model, we used a lightweight DiT with a 1,000-step cosine schedule. To adapt the model for physical surrogates, we integrated heterogeneous boundary conditions and inflow parameters via cross-attention layers. Inference utilized DDIM sampling.

\section{Experiments}\label{sec:experiments}
\subsection{Experimental Setup}

Training data comprised 2,025 two-dimensional CFD simulations using a steady-state solver executed in Simcenter STAR-CCM+ using a RANS k-$\epsilon$ turbulence model for incompressible air flow. The computational domain consisted of a square room measuring 1.5 m × 1.5 m. In each simulation, one 0.1 m wide inlet and one 0.1 m wide outlet were active, located on opposite walls. We evaluated three distinct geometric configurations: an empty room without obstacles, and two configurations containing an obstacle with a diameter of 0.25 m. This obstacle was positioned either squarely in the center of the domain or offset laterally.

For each geometry, we systematically swept 15 inlet positions, 15 outlet positions, and three constant inlet velocities (0.1, 0.2, and 0.3 m/s), resulting in 675 simulations per configuration. The outlet was defined as an atmospheric pressure outlet. The domain was discretized using a polyhedral mesh, with near-wall prism layers applied around the obstacle, while all walls were treated as no-slip wall boundaries. Achieving convergence typically required 750-800 seconds of CPU time per simulation. Following convergence to a steady state, the final flow fields were averaged over the last 100 iterations to mitigate residual numerical fluctuations. Finally, these high-fidelity velocity vectors were interpolated onto a 64×64 uniform grid, with a randomly sampled 10\% of the complete dataset reserved as the held-out test set.


%
%
%

\subsection{Metrics}

We evaluate the predicted velocity field $\hat{\mathbf{u}} = (\hat{u}, \hat{v})$ against the ground truth $\mathbf{u} = ({u}, {v})$ along two axes: field accuracy and flow-structure consistency. All metrics are computed per sample and reported as mean \mbox{$\pm$ standard deviation} over the test set.

\noindent \textit{Field accuracy.}
Following established relative-RMSE metrics in CFD-surrogate evaluation \cite{behrou_physics-informed_2025}, we report the range-normalised RMSE (nRMSE), in which the per-sample RMSE is normalised by the target field's range. As a scale-free complement we report the coefficient of determination ($R^2$), a standard measure of explained variance in indoor-airflow surrogates \cite{reda_rapid_2026, wang_condition_2025}. To capture directional agreement independently of magnitude, we additionally report the cosine similarity between predicted and target vector fields over locations with non-negligible target velocity, which is sensitive to directional disagreement that nRMSE penalises only indirectly \cite{wang_condition_2025}.

\noindent \textit{Flow-structure consistency.}
To probe whether predictions preserve the structural properties of the reference fields, we report two differential diagnostics. We monitor the vorticity as structural quantity relative to the ground truth \cite{li_fourier_2020, kochkov_machine_2021}, to detect over- or under-smoothing of rotational structure. As a complementary diagnostic we monitor the divergence relative to the ground truth; while divergence is often used as a structural constraint in physics-informed learning \cite{mohan_embedding_2020}, we employ it as a passive diagnostic against the target divergence. For both quantities we report the gap between predicted and target magnitudes and the predicted-to-target ratio as a measure of relative structural fidelity.

\subsection{Results}

\subsubsection{Quantitative} 
\leavevmode \\

Table \ref{tab:metrics_small} reports the field-accuracy and flow-structure metrics for the diffusion baseline (chapter \ref{chap:diffusion}) and the two drifting variants on the held-out test set. The diffusion baseline is the strongest model overall, the label-based drifting variant trails it by a small margin, and the spatial drifting variant comes in third with significantly higher variance.


\begin{table}[ht!]
\centering
\caption{Quantitative comparison between the diffusion baseline and our drifting model with label-based and spatial conditioning. The metrics were calculated on obstacle-free test data in pixel space.}
\label{tab:metrics_small}
\small
\begin{tabular}{llccc}
\toprule
\textbf{Category} & \textbf{Metric} & \textbf{Diffusion (baseline)} & \textbf{Drifting (label)} & \textbf{Drifting (spatial)} \\
\midrule
\multirow{3}{*}{\shortstack[l]{Field\\accuracy}} 
& nRMSE (range) $\downarrow$          & $\mathbf{0.0592 \pm 0.0204}$ & $0.0684 \pm 0.0209$ & $0.1076 \pm 0.0453$ \\
& $R^2$ $\uparrow$                   & $\mathbf{0.8476 \pm 0.1228}$ & $0.8019 \pm 0.1152$ & $0.4251 \pm 0.4783$ \\
& Cos. sim. $\uparrow$               & $\mathbf{0.8108 \pm 0.0576}$ & $0.7772 \pm 0.0661$ & $0.7846 \pm 0.1122$ \\
\midrule
\multirow{4}{*}{\shortstack[l]{Flow-struct.\\consistency}} 
& Div. gap (pred vs. target) $\downarrow$ & $0.4503 \pm 0.0849$ & $\mathbf{0.4224 \pm 0.0842}$ & $0.4680 \pm 0.3316$ \\
& Relative divergence $\downarrow$        & $8.0713 \pm 2.1104$ & $\mathbf{7.5579 \pm 1.6388}$ & $8.2887 \pm 5.4017$ \\
& Vort. gap (pred vs. target) $\downarrow$ & $0.2895 \pm 0.1290$ & $\mathbf{0.2634 \pm 0.1312}$ & $0.5519 \pm 0.4618$ \\
& Relative vorticity $\downarrow$         & $1.1635 \pm 0.0820$ & $\mathbf{1.1449 \pm 0.0861}$ & $1.3127 \pm 0.2789$ \\
\bottomrule
\end{tabular}
\end{table}

\noindent \textit{Field accuracy.}
The gap between diffusion and label-based drifting is moderate, supporting our central claim that a single-pass drifting generator can reach competitive accuracy. The spatial variant sacrifices a substantial amount of pixel fidelity, particularly on the nRMSE of $0.108$ versus $0.059$ for the label variant. Cosine similarity shows a smaller gap, indicating that all three models recover the dominant flow direction even where their magnitudes disagree.

\noindent \textit{Flow-structure consistency.}
The label-based drifting model is competitive with the diffusion baseline. It has a smaller divergence gap ($0.422$ vs.\ $0.450$) and a closer relative-vorticity ratio ($1.14$ vs.\ $1.16$). The spatial variant again lags, with a vorticity gap that roughly doubles ($0.552$) and a relative vorticity ratio of $1.31$, indicating systematic over-prediction of rotational structure. The increased standard deviations in the spatial column further suggest that the failure mode is concentrated on a subset of geometries rather than spread uniformly across the test set.

Although the label-based drifting model yields stronger quantitative results in our experiments, its reliance on enumerated training configurations limits its flexibility. We view the spatial encoding as the long-term interface for this class of surrogate, as it fundamentally generalizes to unseen room geometries. We attribute the current performance gap to the foundational nature of this implementation, anticipating that future hyperparameter tuning and encoder refinements will naturally improve accuracy. 


\noindent \textit{Inference speed.} We measured per-sample inference latency on a single NVIDIA A100, timing the full inference pipeline with CUDA events. Under this protocol, the drifting model is two orders of magnitude faster than the diffusion baseline (table \ref{tab:latency}). The low standard deviations confirm that both distributions are tight and outlier-free. The current gap reflects a comparison in a less optimised regime. However, the structural advantages of single-pass inference are clearly visible.

\begin{table}[ht!]
\centering
\caption{Per-sample inference latency on a single NVIDIA A100-SXM4 (80~GB) GPU at batch size~1, over $1000$ runs after $50$ warmup iterations.}
\label{tab:latency}
\small
\begin{tabular}{lccc}
\toprule
\textbf{Model} & \textbf{Conditioning} & \textbf{NFE} & \textbf{Latency (ms)} \\
\midrule
Diffusion & label-based     & $1000$ & $1870.2\phantom{0} \pm 77.1$ \\
Drifting & label-based      & $\phantom{000}1$    & $\phantom{000}\textbf{6.74} \pm \textbf{0.36}$ \\
\bottomrule
\end{tabular}
\end{table}

%

\subsubsection{Qualitative}
\leavevmode \\
%

\begin{figure}[t] 
    \centering
    \includegraphics[width=0.7\textwidth]{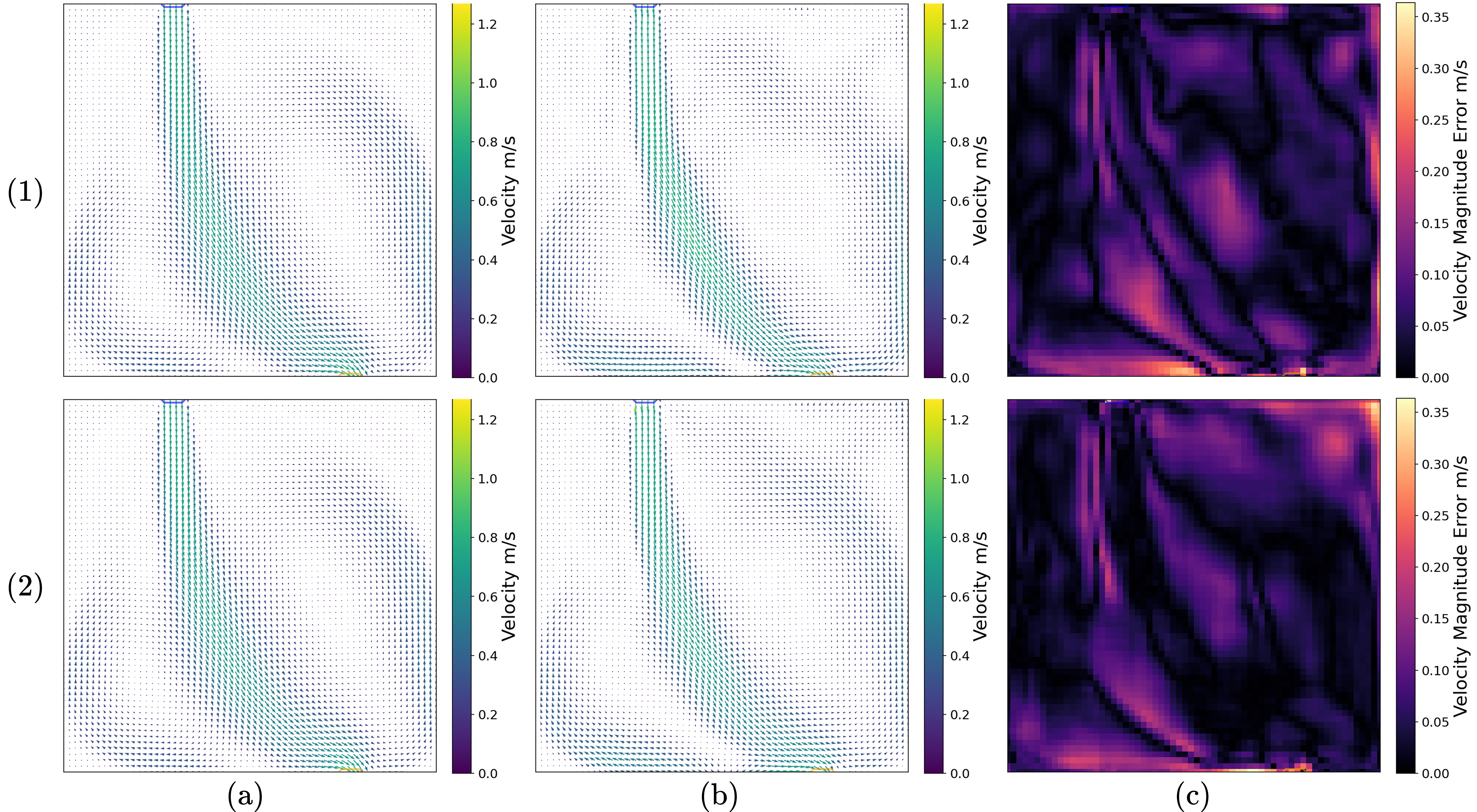}
    \caption{Comparison of predicted velocity fields for the empty-room configuration. (a) is the ground-truth flow field from CFD, (b) the predicted velocity vector fields for (1) label-based drifting model and (2) diffusion baseline, respectively. (c) illustrates the absolute velocity magnitude error compared to the CFD ground truth. Both models were conditioned on an inlet velocity of 0.20 m/s with identical inlet (blue) and outlet (yellow) positions.}
    \label{fig:qualitative_comparison}
\end{figure}

In the empty-room scenario (figure~\ref{fig:qualitative_comparison}), both the diffusion baseline and label-based drifting model successfully recover the primary flow topology, including the dominant jet and corner recirculation zones. Although being a single-pass generator, the drifting model (row 1) achieves high structural fidelity. While diffusion (row 2) shows slightly lower peak errors within the jet, the drifting model’s error distribution is comparably low across most of the domain. This confirms that drifting can match iterative diffusion performance with lower computational cost while preserving essential physics. Both models exhibit peak errors at jet boundaries, likely due to spatial compression within the VAE architecture.

\begin{figure}[t] 
    \centering
    \includegraphics[width=0.99\textwidth]{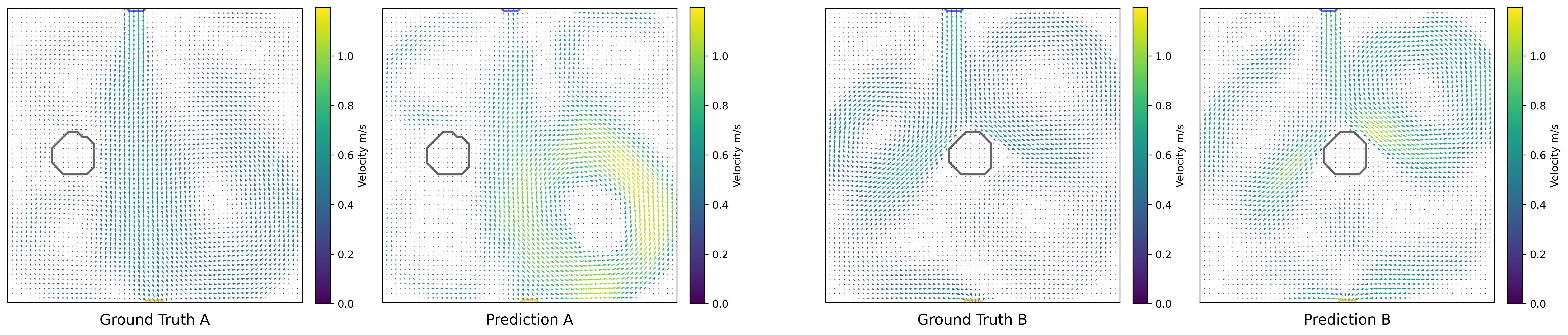}
    \caption{Comparison of two pairs of velocity fields generated by the drifting model with spatial conditioning. Within each pair, the ground-truth simulation is on the left and the model prediction is on the right. Both cases utilize fixed inlet (7) and outlet (8) positions. The left pair features a left-offset obstacle with inlet velocity of 0.3 m/s, while the right pair features a centered obstacle with inlet velocity 0.1 m/s.}
    \label{fig:qualitative_spatial}
\end{figure}

Figure~\ref{fig:qualitative_spatial} illustrates the spatial-conditioning variant's performance with obstacles. While trailing the label-based approach (see table~\ref{tab:metrics_small}), the spatial encoded model captures essential topology, including the primary jet trajectory, flow splitting, and boundary separation around the cylinder. It also reasonably approximates wake and corner recirculation locations. The primary discrepancy is in velocity scaling, where the model produces localized over-predicted speed and rotational structures. These errors directly align with the elevated relative vorticity metrics observed in the quantitative analysis.


\section{Limitations}\label{sec:limits}

Our evaluations focus on 2D steady-state flows on fixed grids to isolate generative fundamentals. However, the dataset's sparsity limits the kernel-based drifting field's potential. Extending to transient 3D flows and multi-state datasets will better leverage the model’s distribution-matching and NFE=1 efficiency.
Performance is currently limited by the VAE's 16× spatial compression; closing the gap to diffusion requires optimizing latent resolution and patch size. Additionally, the spatial conditioning variant's lag likely stems from lossy geometry mask compression, suggesting a need for richer interfaces like cross-attention.
Finally, drifting’s two-order-of-magnitude speed advantage is an inherent structural property, providing predictable, high-speed inference without the quality-speed trade-offs of distilled diffusion. Future benchmarks should include tuned models and direct-regression surrogates to further validate drifting’s utility in steady-state environments.

\section{Conclusion}\label{sec:conlusion}

By adapting a drifting model to 2D steady-state CFD, we achieved performance closely matching a 1000-step diffusion baseline. Although running two orders of magnitude faster, the label-conditioned drifting variant produced an normalised RMSE of $0.068$ compared to the baseline's $0.059$. While diffusion remains more accurate, the narrow gap makes single-step drifting a viable surrogate for cost-constrained applications. Our methodological contributions, latent-space drifting, spatial-scalar conditioning, and label-aware attraction masks, provide a reusable framework for scientific computing surrogates. Future work will target the spatial-conditioning gap and extend the model to transient 3D regimes to solidify drifting as a competitive generative surrogate.


%


\printbibliography[title={References}]

\end{document}